\documentclass[letterpaper]{article} 
\usepackage[draft]{aaai25}  
\usepackage{times}  
\usepackage{helvet}  
\usepackage{courier}  
\usepackage[hyphens]{url}  
\usepackage{graphicx} 
\usepackage{natbib}  
\usepackage{caption} 
\usepackage{amsmath,amssymb}
\usepackage[utf8]{inputenc} 
\usepackage[T1]{fontenc}    
\usepackage{url}            
\usepackage{booktabs}       
\usepackage{amsfonts}       
\usepackage{nicefrac}       
\usepackage{microtype}      
\usepackage{xcolor}         
\usepackage[linesnumbered,ruled,vlined]{algorithm2e}
\usepackage{subcaption,moreverb,marvosym}
\usepackage{algpseudocode}
\def\s{\sigma}
\def\f{\frac}\def\l{\left}\def\r{\right}
\DontPrintSemicolon

\SetKwComment{Comment}{\color{green!50!black}\# }{}

\newcommand{\assign}{\leftarrow}

\newcommand{\FuncCall}[2]{\texttt{\bfseries #1(#2)}}
\SetKwProg{Function}{function}{}{}

\DeclareMathOperator*{\argmax}{arg\,max}
\title{Dartboard: Better RAG using Relevant Information Gain}

\author{
  Marc Pickett,
  Jeremy Hartman,
  Ayan Kumar Bhowmick,
  Raquib Ul Alam,
  Aditya Vempaty
}
\affiliations{
  {\rm Emergence AI} \includegraphics[width=0.02\textwidth]{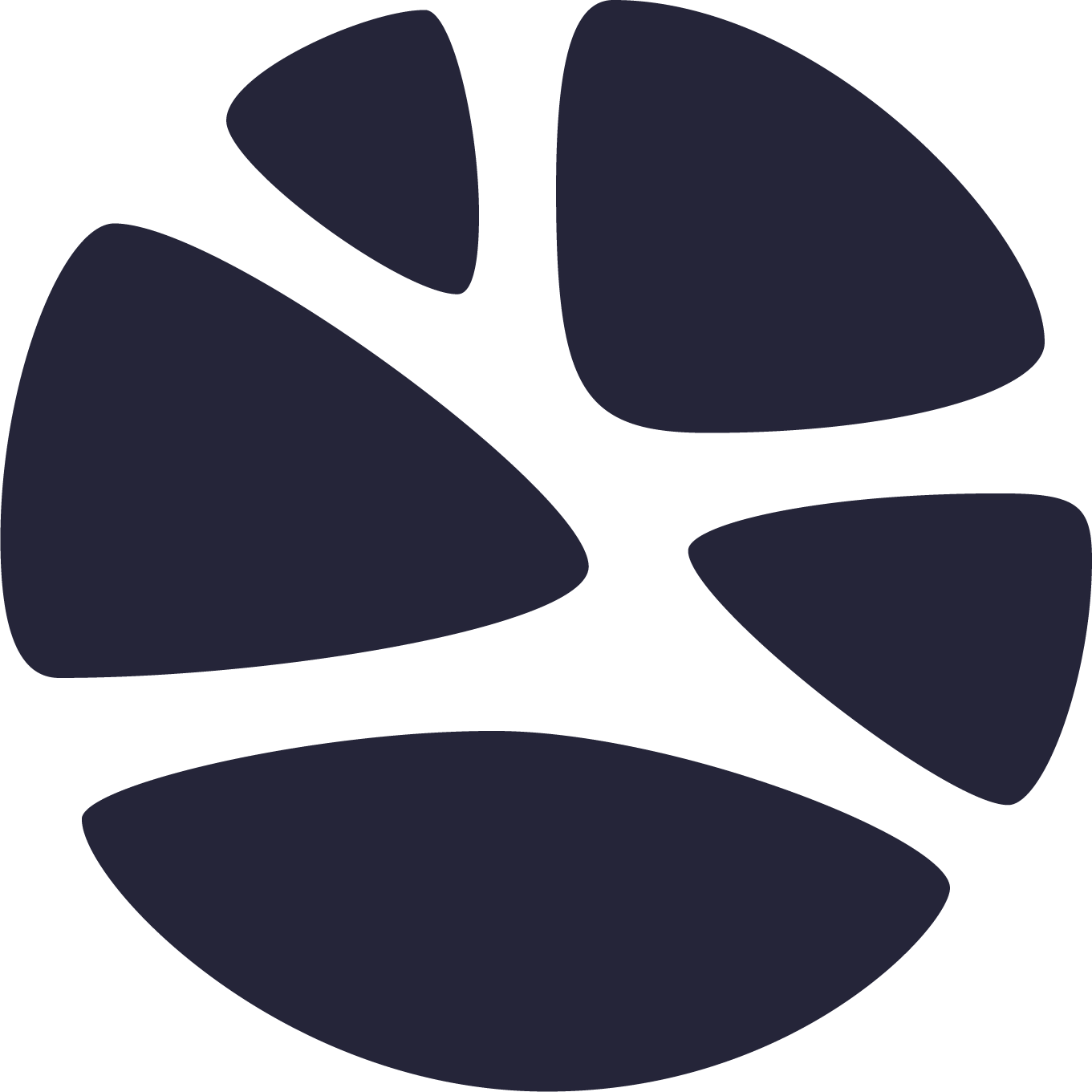}
  \\ \texttt{mpickett@emergence.ai}
}

\begin{document}\maketitle
\begin{abstract} 
  A common way to extend the memory of large language models (LLMs) is by retrieval augmented generation (RAG), which inserts text retrieved
  from a larger memory into an LLM's context window.  However, the context window is typically limited to several thousand tokens, which limits
  the number of retrieved passages that can inform a model's response.  For this reason, it's important to avoid occupying context window space
  with redundant information by ensuring a degree of \emph{diversity} among retrieved passages.
  At the same time, the information should also be \emph{relevant} to the current task. 
  Most prior methods that encourage diversity among retrieved results, such as Maximal Marginal Relevance (MMR), do so by incorporating an
  objective that explicitly trades off diversity and relevance.  We propose a novel simple optimization metric based on \emph{relevant
  information gain}, a probabilistic measure of the \emph{total} information relevant to a query for a set of retrieved results.
  By optimizing this metric, diversity organically emerges from our system.
  When used as a drop-in replacement for the retrieval component of a RAG system, this method yields state-of-the-art performance on question answering tasks from the Retrieval Augmented Generation Benchmark (RGB), outperforming existing metrics that directly optimize for relevance and diversity.
  \footnote{Code is available at \url{https://github.com/EmergenceAI/dartboard}.}
\end{abstract} 

\section{Introduction} 

A limitation of transformer-based Large Language Models (LLMs) is that the number of tokens is bounded by the transformer's context window,
which is typically in the thousands. This is often insufficient for representing large texts, such as novels and corporate documentation. A
common way to mitigate this constraint is via retrieval augmented generation (RAG), in which a relatively small subset of relevant passages are
retrieved from a larger database and inserted into an LLM's context window \cite{gao2024retrievalaugmented}.
Typically, this process involves applying a similarity metric, such as cosine similarity, to (precomputed) embeddings of passages and the embedding
of a query. Using this metric, many systems then use K-nearest-neighbors or a fast approximation with a vector database
such as FAISS \cite{douze2024faiss}.
Importantly, K-nearest-neighbors \cite{bijalwan2014knn} and related methods (such as a cross-encoder reranker \cite{nogueira2020passage}) simply return the highest
\emph{individually} relevant passages, without regard to whether the information in the passages is redundant.
Given the premium value on LLM context-window real estate, it's important to make best use of this limited resource by minimizing redundancy,
while maintaining relevance.

To appreciate the importance of minimizing redundancy in a RAG context, consider a toy database of facts and the two possible sets of
retrieval results in Table \ref{table:sharktar}, for the same query, ``Tell me some facts about sharks.''  Both sets of retrieved results are
highly relevant to the query, but only the second set is diverse enough to support a satisfactory answer.

\begin{table} 
  \footnotesize
  \begin{tabular}{l} 
    \textbf{Shark dataset}
    \\ {\tt Sharks are boneless.}
    \\ {\tt Sharks do not have any bones.}
    \\ {\tt Sharks have no bones.}
    \\ {\tt Sharks have excellent vision.}
    \\ {\tt Sharks are very fierce.}
    \\ {\tt Sharks are apex predators.}
  \end{tabular} 
  \\
  \vspace{10pt}
  \\
  \begin{tabular}{l}
    {\textbf{Query:} Tell me some facts about sharks}\\
    \\
    \textbf{Retrieval results 1:} \\
    {\tt Sharks are boneless.}\\
    {\tt Sharks have no bones.}\\
    {\tt Sharks do not have any bones.}\\
    \\
    \hline\\
    \textbf{Retrieval results 2:} \\
           {\tt Sharks are boneless.} \\
            {\tt Sharks have excellent vision.} \\
            {\tt Sharks are apex predators.}
  \end{tabular} 
  \normalsize
  \caption{A toy database of shark facts (top) and two possible sets of retrieval results for the same query (bottom).}
  \label{table:sharktar}
\end{table} 

A family of methods from the Information Retrieval literature attempts to address the general issue of diversity in retrieved results by
introducing a measure that explicitly balances diversity and relevance \cite{carbonell1998use}.
In this paper, we propose a more principled method, \emph{Dartboard}, that instead seeks to directly accomplish what previous methods are indirectly aiming for - maximize the \emph{total} amount of information relevant for a given query in a set of $k$ results.
The intuition behind \emph{Dartboard} is simple - we assume that one passage is the ``correct'' one for a given query.  Our system is allowed $k$
``guesses'' and it aims to maximize the relevance score of its \emph{most relevant} guess. Since the best guess is not known ahead of time, this score
is weighted by the probability of that guess being the most relevant.
This objective is sufficient to encourage diversity in the guesses.  This is because a redundant guess does little to increase the relevance of
the most relevant guess.

The main contributions of this paper are 3-fold:
\begin{itemize}
\item We introduce the \emph{Dartboard} algorithm, a principled retrieval method based on optimizing a simple metric of total information gain relevant to a given query (\S \ref{section:dartboard}).
\item We demonstrate the effectiveness of \emph{Dartboard} on Retrieval-Augmented Generation Benchmark (RGB)~\cite{chen2023benchmarking}, a closed-domain question answering task.  This benchmark consists of a retrieval component, and an end-to-end question-answering component. We show that the \emph{Dartboard} algorithm, when used as the retrieval component, outperforms all existing baselines at both the component level and at end-to-end level (\S \ref{subsection:results}).
\item We show that instead of directly encouraging diversity, diversity naturally emerges by optimizing this metric (\S \ref{subsection:diversity}).
\end{itemize}

\section{Dartboard} 
\label{section:dartboard}

\begin{figure}[ht]
  \centering
  \includegraphics[width=6cm]{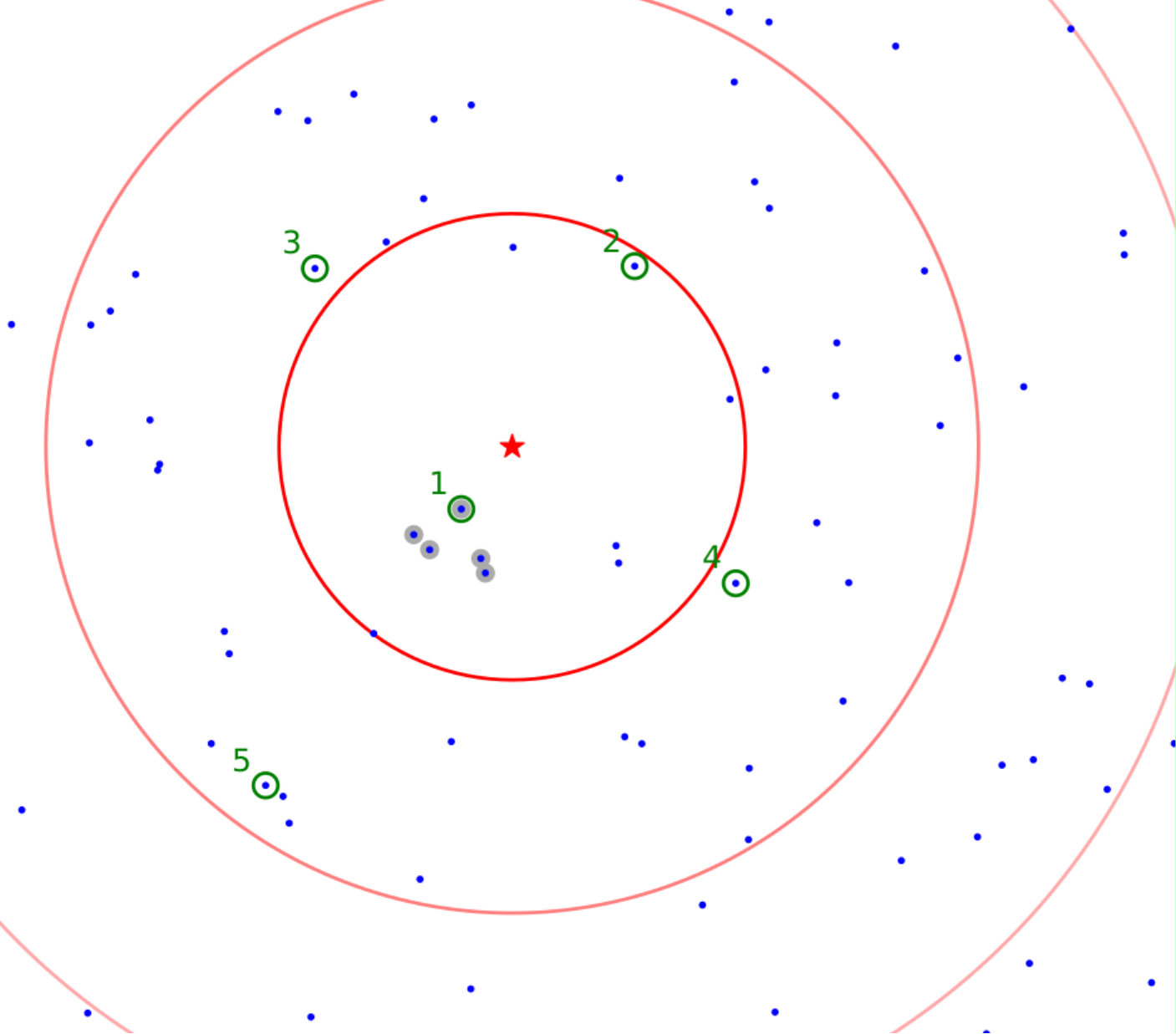}
  \caption{A visualization of \emph{Dartboard}. The \emph{query} is represented by the red star.  All points are represented by blue dots.  The five dots highlighted by grey background are the query's 5 nearest neighbors, while the dots circled in green are the five points selected by the \emph{Dartboard} algorithm (numbered in the order selected by the greedy algorithm). The concentric red circles are spaced at multiples of $\s$, which represents the standard deviation of our uncertainty for the query's accuracy. Note the possible redundancy by naive k-nearest-neighbors, which ignores points above or to the right of the query.}
  \label{figure:dartboard}
\end{figure}

The \emph{Dartboard} algorithm is based on the following analogy illustrated in Figure \ref{figure:dartboard}: Suppose that we have a cooperative
two-player game where a dartboard is covered with a random collection of points.  Player 1 is given one of these points arbitrarily as the
\emph{target}.  Player 1 then throws her dart aiming for the target, and it lands somewhere on the board.  Where it lands is the \emph{query}.  Player
2 sees where Player 1's dart landed (the query), but doesn't know where the actual target is.  Player 2 then picks $k$ of the points on the
board.  The true target is revealed, and the score (which the players are trying to minimize) is the distance from the target to the
\emph{closest} guess.
Note that to minimize the score, Player 2 would not want to put all his guesses right next to each other. Also, Player 2 should take into account
how accurate Player 1's throws are in general. In our implementation, Player 1's accuracy is modeled by a Gaussian distribution with standard deviation
$\s$.

More formally, Player 1 selects a target $T$ from a set of all points $A$ and gives a query $q$. Then Player 2 makes a set of guesses $G
\subseteq A$, resulting in a score $s\l(G, q, A, \s\r)$ which is given as:
\begin{equation} 
  s\l(G, q, A, \s\r) = \sum_{t \in A} P\l(T=t| q, \s\r) \min_{g \in G} D\l(t|g\r)\label{equation:score}
\end{equation} 
where $D$ is a distance function.  For $d$ dimensional vectors, $A \subseteq \mathbb{R}^d$; under some assumptions, we can use a Gaussian kernel
for the distance functions. For example, we can set $P\l(T=t| q, \s\r) = \mathcal{N}\l(q, t, \s\r)$, and we can set $D\l(t|g\r) \propto
-\mathcal{N}\l(t,g,\s\r)$.  Thus, our equation becomes:
\begin{equation} 
  s\l(G, q, A, \s\r) \propto -\sum_{t \in A} \mathcal{N}\l(q, t, \s\r) \max_{g \in G} \mathcal{N}\l(t,g,\s\r)
  \label{equation:cosine}
\end{equation} 


\subsection{The Dartboard Algorithm} 

The \emph{Dartboard} Algorithm aims to maximize Equation \ref{equation:cosine} given a distance metric.  In practice, we can greedily build our set
$G$, which works well as it saves us combinatorial search, and allows reuse of previous answers (since the top-$k$ results are a subset of the
top-$k+1$ results).  We begin by ranking top-$k$ passages $A'$ from our initial dataset of passages $A$ using $K$-nearest-neighbors based on cosine similarity.
We use a linear search, but sub-linear methods such as FAISS \cite{douze2024faiss} could also be used for this initial ranking. Our search
is a simple greedy optimization method with two changes - (a) we stay in log space to avoid numerical underflow, and (b) we reuse the results
($maxes$) from previous loops to avoid recomputing the maximums. The detailed algorithm is given in Algorithm \ref{algorithm:dartboard} in
Appendix \ref{sec:algorithm}. In Appendix \ref{sec:reranker}, we also show how to adapt \emph{Dartboard} to use a cross-encoder based reranker (resulting in two methods called \emph{Dartboard crosscoder} and \emph{Dartboard hybrid}), and Appendix \ref{subsection:generalizes} shows that \emph{Dartboard} generalizes KNN and MMR retrieval algorithms~\cite{onal2015utilizing}.

\section{Experiments} 

We tested \emph{Dartboard} on benchmark datasets from \cite{chen2023benchmarking}, from which we used two types of closed-domain question answering.
In the \emph{simple question answering} case, a query is answerable from a single passage retrieved from the corpus. For example, consider the query {\tt When is the premiere of `Carole King \& James Taylor: Just Call Out My Name'?}. On the other hand, in the \emph{information integration} case, a query would require multiple
passages to be retrieved to answer the query. For example, consider the query {\tt Who is the director of `Carole King \& James Taylor: Just Call Out My Name' and when is its premiere?}. We modified this benchmark for our setup in the following way. The original benchmark contains ``positive'' and ``negative'' labeled
passages for each query. The positive passages are useful for answering, while the negative ones are related but ineffective in answering the query. Since we are interested in the retrieval component of this task, we merged the positive and negative passages for all queries into a single
collection of $11,641$ passages for the $300$ \emph{simple question answering} test cases and $5,701$ passages for the $100$ \emph{information integration} test cases.
The evaluation is otherwise identical apart from the retrieval component.  Note that the innovation of \emph{Dartboard} is solely on the retrieval component.
Therefore, we keep the rest of the RAG pipeline fixed. In particular, we do not modify the prompting of LLMs or try to
optimize passage embeddings.

Given a query and the full set of thousands of passage embeddings, we measured both a direct retrieval score and the overall end-to-end performance of the
system with the only change being the retrieval algorithm.  For the direct retrieval score, we computed the Normalized Discounted Cumulative
Gain (NDCG) score~\cite{wang2013theoretical} on retrieving any one of the ``positive'' passages relevant to a specific query. In the information integration case, the positive
passages were split into positive ones for each component of the question.  Therefore, in this case, we calculated the NDCG score for retrieving at least one positive passage for \emph{each} component of the query.
For the end-to-end score, given an LLM's response to the query (generated from retrieved passages), we use the same evaluation as \cite{chen2023benchmarking}, which
does a string match of the response on a set of correct answers, marking each response as either correct or incorrect.

Some of the methods (described in Appendix \ref{sec:baselines}), including \emph{Dartboard}, have tunable parameters. For instance, Maximal Marginal Relevance
(MMR) has a diversity parameter that varies from 0 to 1. We performed a grid search over these parameters, reporting the best results for each
method.


\subsection{Results} 
\label{subsection:results}



From the results shown in Table \ref{tab:results}, we observe that \emph{Dartboard} outperforms all state-of-the-art methods in terms of all metrics across all the tasks.

\begin{table}[ht]
  \centering
  \footnotesize
  \begin{tabular}{lrrrr}
    \toprule
    & \multicolumn{2}{c}{Simple} & \multicolumn{2}{c}{Integrated} \\
    &                                       QA &       NDCG &   QA & NDCG \\
    \midrule
    \emph{Oracle}                    &     89.3\% &     1.000 &     36\% &     .826 \\
    D-H ({\bf ours})    &{\bf 85.6\%}&     0.973 &     41\% &{\bf .609}\\
    D-CC ({\bf ours})        &     84.3\% &     0.971 &{\bf 42\%}&     .595 \\
    D-CS ({\bf ours})           &     83.0\% &{\bf 0.975}&     36\% &     .545 \\
    MMR Crosscoder                   &     84.3\% &     0.971 &     40\% &     .598 \\
    MMR Cossim                       &     81.0\% &     0.974 &     36\% &     .541 \\
    KNN Crosscoder                   &     84.3\% &     0.968 &     36\% &     .580 \\
    KNN Cossim                       &     80.0\% &     0.973 &     25\% &     .514 \\
    Empty                            &      3.3\% &     0.000 &      3\% &     .000 \\
    Random                           &      3.3\% &     0.044 &      2\% &     .028 \\
    \bottomrule
  \end{tabular}
  \normalsize
  \caption{Results for the \emph{Dartboard} retrieval system on the QA benchmarks using $k=5$.  For methods with tunable parameters (\emph{Dartboard} and MMR), the best score over a parameter sweep is reported.}
  \label{tab:results}
\end{table}

\begin{figure}[ht]
  \centering
  \includegraphics[width=8cm]{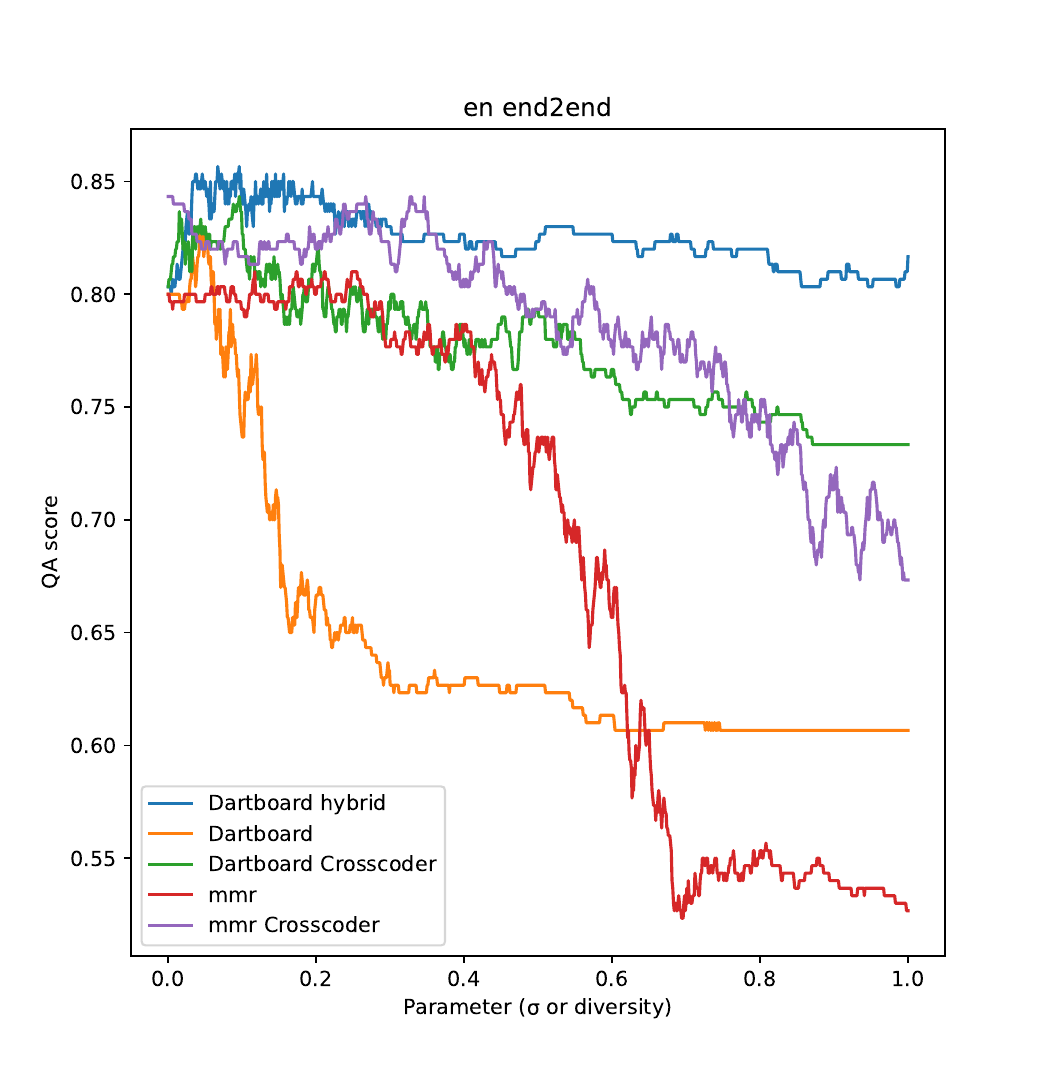}
  \caption{Performance on end-to-end QA task (simple) as parameters vary.  For \emph{Dartboard}, we show its performance as $\s$ varies.  For MMR, we
    show its performance as the diversity parameter varies.}
  \label{figure:parameters}
\end{figure}

Figure \ref{figure:parameters} shows the performance of different retrieval methods on the end-to-end QA task (simple) as the parameters
vary. Although \emph{Dartboard Crosscoder (D-CC)} and \emph{Dartboard hybrid (D-H)} are fairly robust to a range of $\s$ values, the best
performance is achieved for \emph{Dartboard hybrid} with $\s=0.096$ (See Appendix \ref{sec:baselines} for baselines).

\section{Related Work} 
MMR retrieves documents~\cite{carbonell1998use} that are both relevant to the query and dissimilar to previously
retrieved documents. It combines a relevance score (e.g., from BM25) with a novelty score that penalizes documents similar to those already
retrieved. It have been used extensively for building recommendation systems~\cite{xia2015learning,wu2023maximal} as well as for summarization
tasks~\cite{agarwal2022extractive,adams2022combining}. However, MMR suffers from few limitations.  First is that MMR requires the \emph{diversity} parameter to control the balance between relevance and novelty. This parameter is often dataset-specific and requires careful tuning,
making it impractical for real-world applications. Second is that MMR can favor exact duplicates of previously retrieved documents as they
retain a high relevance score while minimally impacting the average novelty score (See Appendix \ref{sec:duplicates}).

KNN retrieves documents based on their similarity to a query embedding~\cite{dharani2013content,bijalwan2014knn}. While efficient, KNN often
suffers from redundancy as nearby documents in the embedding space tend to be semantically similar~\cite{taunk2019brief}. This can lead to a
retrieved set dominated by passages conveying the same information with slight variations.

Several recent works have explored incorporating diversity objectives into retrieval
models~\cite{angel2011efficient,li2015diversity,fromm2021diversity}. These approaches often involve complex optimization functions or require
additional training data for diversity estimation.
For example, Learning-to-Rank with Diversity methods leverage learning-to-rank frameworks that incorporate diversity objectives directly into
the ranking function. This allows for the optimization of both relevance and diversity during the ranking process. However, these approaches
often require large amounts of labeled training data for diversity, which can be expensive and time-consuming to
obtain~\cite{wasilewski2016incorporating,yan2021diversification}.
Bandit-based approaches model document selection as a multi-armed bandit problem~\cite{hofmann2011contextual,wang2021interactive}. The model
explores different retrieval strategies and receives feedback based on the relevance and diversity of the retrieved passages. These approaches
can be effective but can be computationally expensive for large-scale retrieval tasks.

RAG models have also been extended to incorporate diversity objectives.  For example, RAG with Dense Passage Retrieval retrieves a large number
of candidate passages~\cite{cuconasu2024power,reichman2024retrieval,siriwardhana2023improving}. It then employs a two-stage selection process:
first selecting a diverse subset based on novelty scores, then selecting the most relevant passages from this subset.  While
effective, this approach requires careful tuning of the selection thresholds.

\section{Discussion} 

In this paper, we introduce \emph{Dartboard}, a principled retrieval algorithm that implicitly encourages diversity of retrieved passages by optimizing
for relevant information gain.  We demonstrate that \emph{Dartboard} outperforms existing state-of-the-art retrieval algorithms on both retrieval and end-to-end QA tasks.
We view this work as an initial step for a more general line of work that optimizes information gain during retrieval, especially in the context
of RAG systems.  In future work, we plan to investigate \emph{Dartboard} for other retrieval tasks, such as suggestion generation (see Appendix
\ref{sec:generative}).

\section{Limitations} 
We have not done a systematic investigation of the run time of \emph{Dartboard}. In the worst case scenario, \emph{Dartboard} is quadratic in the number of ranked
passages. However, in practice, \emph{Dartboard hybrid} typically runs in a fraction of a second for ranking (based on cosine-similarity with query) a set of $100$
passages (note that a full cross-encoder based MMR/\emph{Dartboard} needs to run the cross-encoder $10,000$ times, and can take several seconds). This
retrieval time is minimal compared to the time required for a LLM to process the retrieved passages and generate an answer.

Our experimental results are limited to a single benchmark and a single LLM i.e. ChatGLM~\cite{hou2024chatglm}. It remains to be seen whether our results would generalize to other benchmarks and LLMs. We plan to investigate this in future work.

One shortcoming of our method (also shared by MMR) is that it requires a hyperparameter that affects how much diversity is encouraged. While we show that \emph{Dartboard} is robust to the choice of this hyperparameter, it would be ideal to have a method that does not require manual
tuning. As part of future work, we plan to investigate methods that automatically adapt to the context of the query. For example, the hyperparameter could be set based on a held-out validation set.

Another topic for future work is to investigate if it is also possible for $\s$ to vary depending on the type of query. For example, a query like
``Tell me facts about The Beatles'' would warrant a broader range of passages than a query like ``Tell me facts about George Harrison''.

Another shortcoming of our approach is that our benchmarking criteria is limited in terms of the evaluation protocol we are using. Our evaluation is based on an exact string match of the
output answer generated from the LLM with a set of possible answers. For example, for one question, the generated output answer is considered correct if it contains the exact string {\tt
  `January 2 2022'}, {\tt `Jan 2, 2022'}, etc., but would be considered incorrect if it only contains {\tt `January 2nd, 2022'}.  However, we left
the benchmark as is (modulo our modifications mentioned above) so that our method is easily comparable to that of others.

Finally, though the initial cosine similarity based proposed \emph{Dartboard} method is principled, the hybrid variation of \emph{Dartboard} is not that principled. This is because it tries to compare logits from a cross-encoder with the cosine similarity of a different embedding model, similar to comparing apples with oranges, though it seems to work well as seen in
our presented empirical results.

\bibliography{dartboard}
\appendix
\section{Appendix}
\label{sec:appendix}

\subsection{Dartboard Algorithm Details} 
\label{sec:algorithm}
The full algorithm for Dartboard is described in Algorithm \ref{algorithm:dartboard}.

\begin{algorithm}
  \caption{Dartboard}\label{algorithm:dartboard}
  \Comment{Natural log of Gaussian pdf.}
  \Function{LogNorm($\mu$, $\s$)}{
    \Return{$\ln\l(\s\r) - \f{1}{2} \ln\l(2 \pi\r) - \f{\mu^2}{2 \s^2}$}\;
  }
  \Comment{$q$: the query.}
  \Comment{$A$: set of all points.}
  \Comment{$K$: number of points to triage.}
  \Comment{$k$: number of points to return.}
  \Comment{$\s$: The standard deviation of the Gaussians.  A measure of spread.}
  \Function{Dartboard($q$, $A$, $K$, $k$, $\s$)}{
    \Comment{Triage and get the distances.}
    $A', ids \assign \FuncCall{KNN}{$q$, $A$, $K$}$ \Comment{Triage using KNN.}\;
    $D \assign \FuncCall{dists}{$A'$, $A'$, $\s$}$ \Comment{$K$x$K$ distance matrix.}\;
    $Q \assign \FuncCall{dists}{$q$, $A'$, $\s$}$ \Comment{Distance from each $a \in A'$ to $q$.}\;
    \Comment{Work in log space for numerical stability.}
    \Comment{Note that $D$ and $Q$ are now log probabilities, not distances.}
    $D \assign \FuncCall{LogNorm}{$D$, $\s$}$\;
    $Q \assign \FuncCall{LogNorm}{$Q$, $\s$}$\;
    \Comment{Greedily seed and search.}
    \Comment{We only track the last addition's contribution.}
    $m \assign \argmax_i\l(Q\r)$\;
    $maxes \assign D_m$\;
    $ret \assign \l[ids_m\r]$\;
    \Comment{Incrementally add until we have $k$ elements.}
    \While{$\l|ret\r| < k$}{
      $newmax \assign \max\l(maxes, D\r)$\;
      $scores \assign \FuncCall{LogSumExp}{$newmax + Q$}$\;
      \Comment{Get the best candidate.}
      $m \assign \argmax_i\l(scores\r)$\;
      $maxes \assign newmax_m$\;
      $ret \assign \FuncCall{append}{$ret$, $ids_m$}$\;
    }
    \Return{$ret$}\;
  }
\end{algorithm}

\subsection{Baselines} 
\label{sec:baselines}

In this section, we briefly describe the different variations of \emph{Dartboard} as well as the competing retrieval methods that we use to compare the performance of \emph{Dartboard} in Table~\ref{tab:results} in the main paper. All methods that rely on using the cross-encoder first use KNN to retrieve the top $100$ passages.
\begin{itemize}
\item{\textbf{Dartboard cossim (D-CS):}} This is the variation of the proposed \emph{Dartboard} method that relies on using cosine similarity for ranking passages.
\item{\textbf{Dartboard crosscoder (D-CC):}} This is the variation of the proposed \emph{Dartboard} method that relies on using cross-encoder based similarity.
\item{\textbf{Dartboard hybrid (D-H):}} This is the variation of the proposed \emph{Dartboard} method that relies on using cross-encoder for the Gaussian kernel $\mathcal{N}\l(q,t,\s\r)$ and cosine similarity for the Gaussian kernel $\mathcal{N}\l(t,g,\s\r)$.
\item{\textbf{KNN cossim:}} This is the variation of K-nearest neighbors algorithm that relies on using using cosine similarity.
\item{\textbf{KNN crosscoder:}} This is the variation of K-nearest neighbors algorithm that relies on using cross-encoder similarity.
\item{\textbf{MMR cossim:}} This is the variation of the Maximal Marginal Relevance method that relies on using cosine similarity.
\item{\textbf{MMR crosscoder:}} This is the variation of the Maximal Marginal Relevance method that relies on using cross-encoder similarity.
\item{\textbf{Empty:}} This is a method that involves no retrieval step but uses just the LLM to generate the answer for a given query.
\item{\textbf{Oracle:}}  This method retrieves only the ``positive'' labeled passages. For the information integration case, we retrieve positive passages for each component of the query up to $k$. If the number of positive passages is less than $k$, we use the negative passages to fill in the rest.
\item{\textbf{Random:}} This method randomly retrieves $k$ passages from the full passage set.
\end{itemize}

\subsection{Modification for cross-encoder based reranker} 
\label{sec:reranker}
Cross-encoder-based reranking has been shown to outperform embedding-based approaches such as cosine similarity \cite{nogueira2020passage},
as it uses the full computational power of a transformer model, rather than being limited to simple vector operations. We have proposed two variations of \emph{Dartboard}, namely \emph{Dartboard Crosscoder} and \emph{Dartboard Hybrid}, based on how we compute the cross-encoder scores for the Gaussian kernels in Equation \ref{equation:cosine} given in the main paper.
For the \emph{Dartboard Crosscoder} variation, we use the cross-encoder score $C\l(q,t\r)$ before computing the Gaussian kernel for
both $\mathcal{N}\l(q, t, \s\r)$ and $\mathcal{N}\l(t,g,\s\r)$ in Equation \ref{equation:cosine}.  Note that the cross-encoder score is
asymmetric, so we simply average the two possible ways to compute the cross-encoder score for $\mathcal{N}\l(t,g,\s\r)$, i.e., $\f{1}{2}\l(C\l(t,g\r) + C\l(g,t\r)\r)$.  For
$\mathcal{N}\l(q, t, \s\r)$, we are only interested in the likelihood of $t$ given $q$, so we only use the cross-encoder score $C\l(q,t\r)$.

However, the cross-encoder is computationally expensive to run for $k^2$ pairs. Hence, we rely on the \emph{Dartboard-Hybrid} variation wherein we use the cross-encoder score only for
the Gaussian kernel $\mathcal{N}\l(q, t, \s\r)$ whereas we use cosine similarity for the Gaussian kernel $\mathcal{N}\l(t,g,\s\r)$.

\subsection{Dartboard generalizes KNN and MMR}
\label{subsection:generalizes}

The \emph{Dartboard} algorithm can be viewed as a generalization of the traditional retrieval algorithms, KNN and MMR. In order to verify this claim, let us look at the score presented in
Equation~\ref{equation:score} in the main paper. When the Player 1 has a perfect aim, or in other words, $\sigma \to 0$, $P(T=t|q,\sigma)$ tends to a point mass
distribution such that $t=q$, and hence the score becomes
\begin{equation}
    s\l(G, q, A, \s\r) \to \min_{g \in G} D\l(q|g\r)
\end{equation}
where $D$ is the distance function as before. If the chosen distance function is proportional to the similarity measure, this is nothing but the
KNN algorithm. On the other hand, when the chosen distance function is the weighted sum of the similarity between query and guess, and dissimilarity between
current guess and past guesses, it reduces to the MMR algorithm.

\subsection{Dartboard inherently promotes diversity} 
\label{subsection:diversity}

\begin{figure}[ht]
  \centering
  \includegraphics[width=\columnwidth]{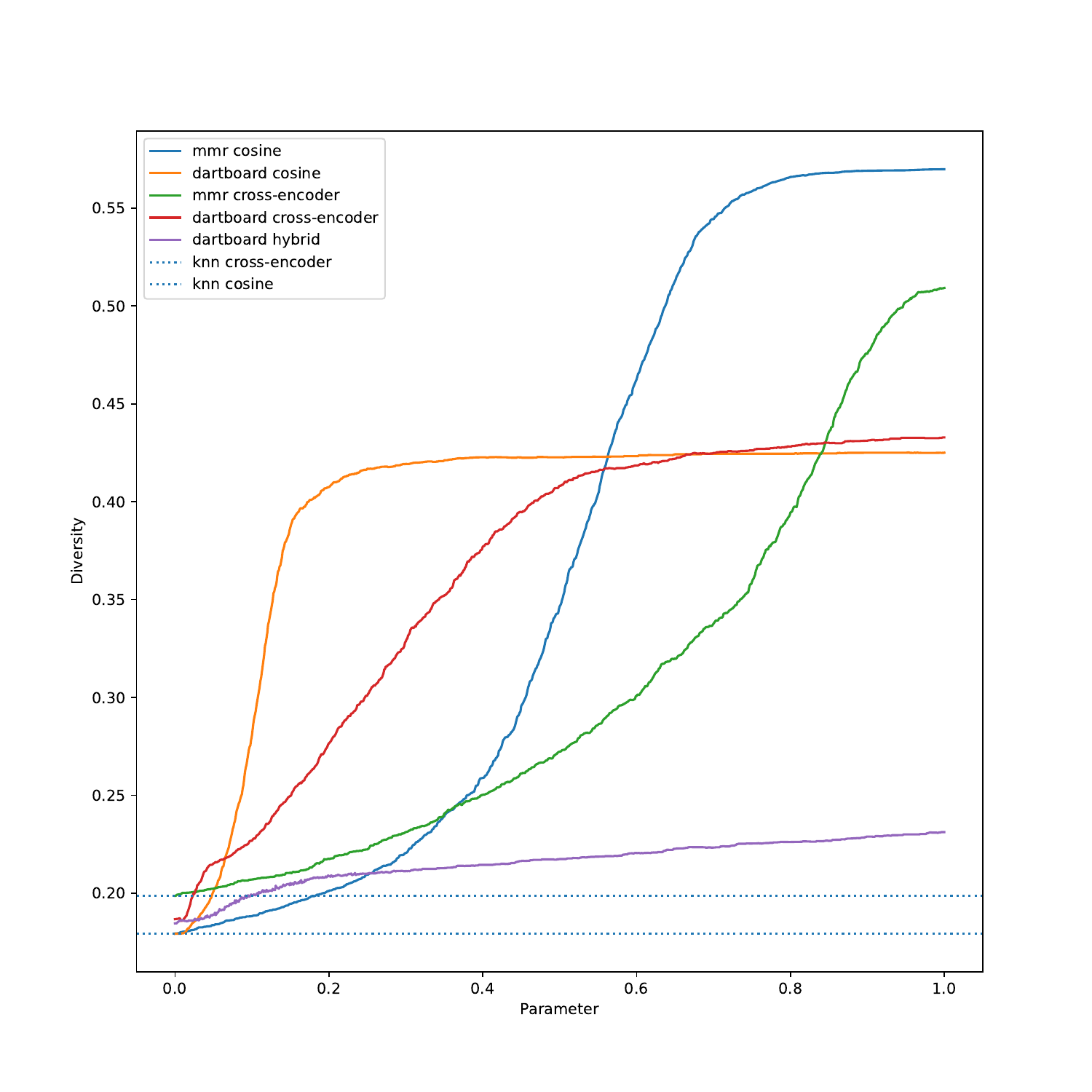}
  \caption{We show the diversity in the set of retrieved passages from RGB for both \emph{Dartboard} and MMR (for $k=5$), where diversity is one minus the average cosine similarity between pairs of retrieved passages. For both MMR and \emph{Dartboard}, diversity increases as the value of the parameters ($\s$ and \emph{diversity} for
    \emph{Dartboard} and MMR respectively) increases.}
  \label{figure:diversity}
\end{figure}

In Figure \ref{figure:diversity}, we show the diversity of the retrieved passages from RGB for both \emph{Dartboard} and MMR, measured as one minus the average cosine similarity between pairs of retrieved passages. While MMR explicitly
encourages diversity, Dartboard does not. However, we observe from the figure that as the parameter $\s$ increases, the diversity of the retrieved passages also
increases. This implies that by optimizing the relevant information gain metric, \emph{Dartboard} inherently ensures diversity in the set of retrieved passages.

\subsection{Example of a generative use of Dartboard} 
\label{sec:generative}
Below is an example of the set of retrieved passages for a query that shows that the passages retrieved by \emph{Dartboard} are highly diverse compared to those retrieved by KNN which has high redundancy, if we consider the cross-encoder based variations:

\footnotesize
\begin{verbatim}
Query: Do you want to watch soccer?

Candidates:
 1: Absolutely!
 2: Affirmative!
 3: I don't know!
 4: I'd love to!
 5: Maybe later.
 6: Maybe!
 7: Maybe...
 8: No thanks.
 9: No way!
10: No, I don't wanna do dat.
11: No, thank you!
12: No, thank you.
13: Not right now.
14: Not today.
15: Perhaps..
16: Sure!
17: Yeah!
18: Yes!
19: Yes, please can we?
20: Yes, please!
21: Yes, please.
22: Yes, we ought to!
23: Yes, we shall!
24: Yes, we should!

KNN crosscoder:
  18: Yes!
  21: Yes, please.
  20: Yes, please!

Dartboard crosscoder:
  18: Yes!
   7: Maybe...
  12: No, thank you.
\end{verbatim}
\normalsize

\subsection{Dartboard does not allow for the possibility of exact duplicates} 
\label{sec:duplicates}

The ``max'' in Equation \ref{equation:cosine} given in the main paper ensures that the same vector (passage) is not selected twice (unless all non-duplicate/unique passages have been exhausted) in case of \emph{Dartboard}.  This is in contrast to MMR, which can select the same vector (passage).

Here is an example where MMR produces exact duplicates. Consider the scenario when our passage database consists of the vectors $\{(2, 1), (2, 1), (1, 2), (0, 1)\}$ (with a duplicate $(2, 1)$). Now if we use cosine
similarity based scoring, and set diversity to $.5$ for $k=3$ in case of MMR, the bag that maximizes the score for probe $(2, 1)$ for MMR is
$\{(0, 1), (2, 1), (2, 1)\}$, which has an exact duplicate passage vector $(2, 1)$.
This verifies that MMR can allow for exact duplicates, which can increase the MMR score because it decreases the average distance to the query, while (possibly) only
marginally decreasing the diversity.

On the contrary, in case of \emph{Dartboard}, an exact duplicate passage vector will add zero information i.e. it would not increase the chances of hitting the target. So it will not be selected for retrieval until all other non-duplicate options are exhausted.

\subsection{More results} 

\begin{figure}[ht]
  \centering
  \includegraphics[width=\columnwidth]{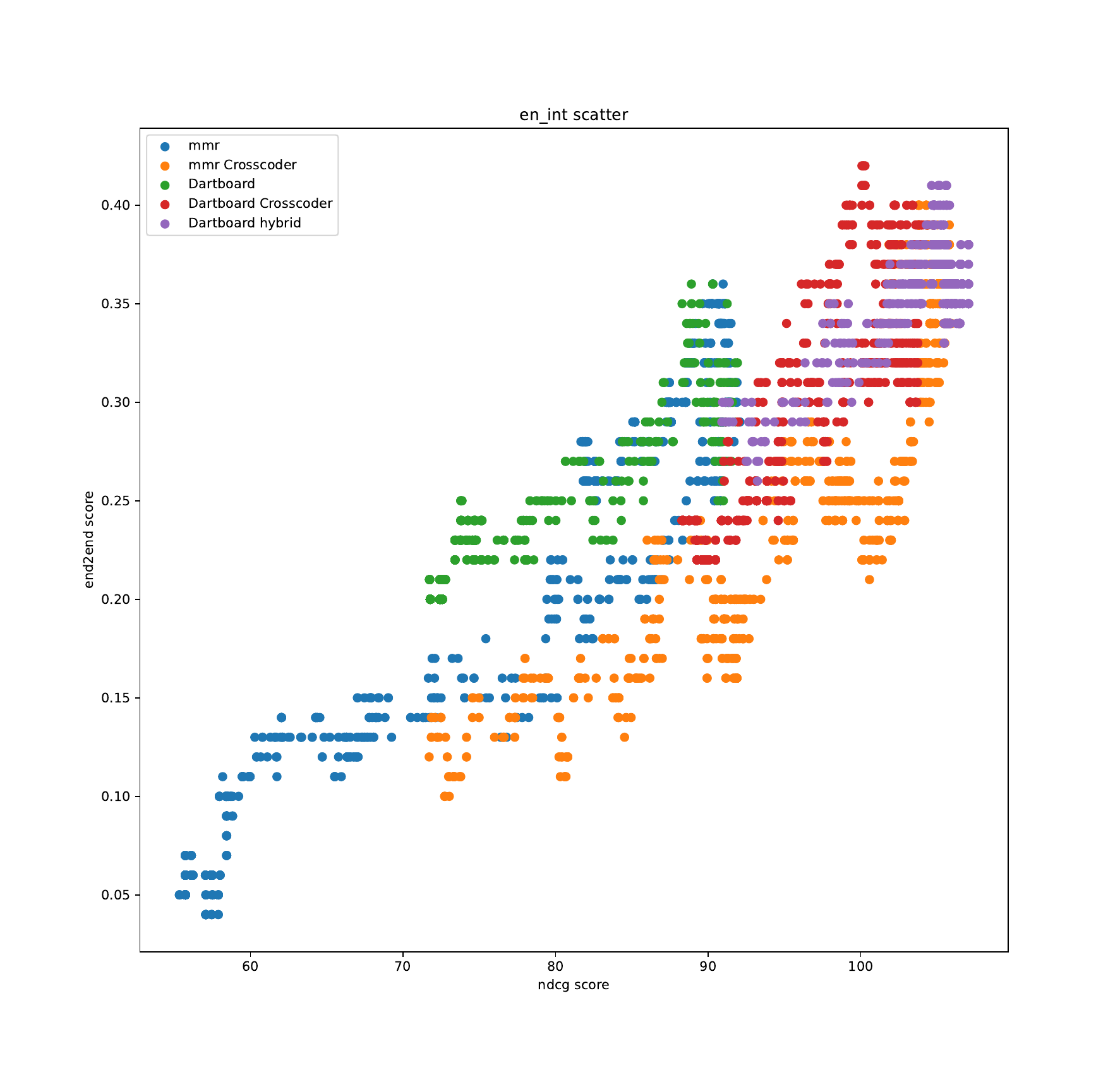}
  \caption{Scatter plot of NDCG score and final end-to-end performance on the QA task. The best performing methods are in the upper right hand side of the plot.}
  \label{figure:scatter}
\end{figure}

In Figure \ref{figure:scatter}, we show the relation between NDCG score and final end-to-end performance on the question answering (QA) task.

\end{document}